# DR²L: Surfacing Corner Cases to Robustify Autonomous Driving via Domain Randomization Reinforcement Learning

Haoyi Niu
Department of Automation/Tsinghua University
Beijing, China
niuhy18@mails.tsinghua.edu.cn

Jianming Hu*
Department of Automation/Tsinghua University
Beijing, China
hujm@mail.tsinghua.edu.cn

Zheyu Cui
Department of Automation/Tsinghua University
Beijing, China
czyfaiz@foxmail.com

Yi Zhang
Department of Automation/Tsinghua University
Beijing, China
zhyi@mail.tsinghua.edu.cn

## ABSTRACT

How to explore corner cases as efficiently and thoroughly as possible has long been one of the top concerns in the context of deep reinforcement learning (DeepRL) autonomous driving. Training with simulated data is less costly and dangerous than utilizing real-world data, but the inconsistency of parameter distribution and the incorrect system modeling in simulators always lead to an inevitable Sim2real gap, which probably accounts for the underperformance in novel, anomalous and risky cases that simulators can hardly generate. Domain Randomization(DR) is a methodology that can bridge this gap with little or no real-world data. Consequently, in this research, an adversarial model is put forward to robustify DeepRL-based autonomous vehicles trained in simulation to gradually surfacing harder events, so that the models could readily transfer to the real world.

## 1 Introduction

Having witnessed a multitude of DeepRL-based autonomous driving techniques[1] rapidly emerging with outperformance in the simulated domain, we cannot help addressing the problem of whether these models could showcase comparable performance in the physical world. From previous work came a suite of discoveries answering this question, autonomous vehicles possibly eliminate normal accident cases such as rear-end collisions but hardly cover the edge cases, and they even produce whole new unexpected cases[2] in the real world. Additionally, confirming the reliability of these vehicles trained in simulators requires driving for hundreds of millions of miles[3]. Brooks[4] expresses his great concern about the difficulties to simulate the actual dynamics of the real world and points out that programs that work well on simulated robots always completely fail on real robots(including robot-like vehicles) because of the differences in real-world sensing and actuation. Therefore, models trained in simulators would surely face the Sim2real gap[5].

To narrow the gap, the most direct approach is apparently sampling in the real world. However, DeepRL algorithms are data-hungry and a large body of interactions with the real world is absolutely unsafe for undertrained autonomous vehicles. On the contrary, simulated data are cheap, safe, and scalable but scarcely match the real-data distribution and customarily miss the corner cases. Domain Randomization(DR), a simple but powerful approach that could transfer the simulated training model to the real world, has gradually shouldered the responsibility in real robot manipulation tasks[6]. In the following sections, we would like to implement Domain Randomization Reinforcement Learning(DR²L) via which autonomous vehicles could be robust to corner cases with little preliminary knowledge about the real world. In this research, we validate that vehicles trained in the DR²L model have a stronger generalization and better performance in efficiency and safety.

The remainder of this paper is structured as follows: In Section II we elaborate on different approaches to close the Sim2real gap and handle the edge case problems, and also demonstrate the validity of DR. A detailed description about how we model the DR²L algorithm and how it performs follows in Section III and Section IV respectively. Finally, we draw our conclusion and put forward considerable future work in Section V.

## 2 Related Work

To cover the corner cases that simulators always neglect, a couple of methods have been devised. Here are a couple of key concepts that should be stated in advance: (1)Source Domain: (i.e. simulator) the domain where training occurs; (2)Target

Domain: (i.e. real world) the domain that we'd like to transfer the simulated model to.

## 2.1 System Identification

System Identification (SI) is to model a simulated system, which needs fine calibration to match the real-world data distribution[7]. The departure point, approximating Source Domain to Target Domain, sounds reliable and accurate; nevertheless, calibration is costly and, to some extent, unrealistic, because physical parameters(e.g. temperature, humidity, friction or wear-and-tear of vehicles mechanical components) in the driving environment vary drastically, and what's more, it is dangerous for autonomous vehicles to access the real data distribution for calibration. To this end, Yu[8] frames the simulator system as two parts: a universal policy and a function for Online System Identification(OSI), which uses the recent state and action history of the system to predict the dynamics model parameters. Undoubtedly, OSI is safer, but the parameter prediction idea is a compromise towards the high-fidelity of classical SI.

## 2.2 Domain Adaptation

Domain Adaptation(DA) is a group of transfer learning methods that map the data distribution in Source Domain to the one in Target Domain. Among DA approaches, GAN[9][10] is a commonly-used tool to adapt simulated images to look like as if captured from the Target Domain. Accordingly, DA with GAN is appropriate to end-to-end vision-based autonomous driving. The only drawback is the requirement for innumerous real data from Target Domain.

## 2.3 Domain Randomization

Domain Randomization(DR) is a relatively simpler idea that works just by randomizing the dynamics properties of the Source Domain during training. Assuming the state of vehicles could be described as an N-dimensional vector, we sample a set of N randomization parameters $\xi$ from a randomization space $\Xi \subset \mathbb{R}^N$ in the Source Domain $e_\xi$. The training policy refers to $\pi_\theta$. $\tau_\xi$ is a trajectory collected in source domain randomized with $\xi$. The policy parameter $\theta$ is trained to maximize the expected reward average across a distribution of configurations:

$$\theta^* = \mathop{\mathrm{argmax}}_{\theta} \mathbb{E}_{\xi \sim \Xi}\left[\mathbb{E}_{\pi_\theta, \tau \sim e_\xi}[R(\tau)]\right] \quad (1)$$

Peng et al.[11] master a good understanding of the essence: Discrepancies between the Source and Target Domains are modeled as variability in the Source Domain.

### 2.3.1 *Uniform Domain Randomization*

Uniform Domain Randomization(UniformDR), namely the original form of DR, bound each randomization parameter $\xi_i$ in a range where they are uniformly sampled. Tobin et al. [11] randomize the scene properties such as textures, colors, camera position, and orientation in the context of robot manipulation problems. In addition to this type of *Visual Domain Randomization*, physical dynamics in the simulator could also be randomized, namely *Dynamics Domain Randomization*, which focuses on the robustness to uncertainty in system's dynamics such as friction, discretization timestep, and random forces applied to the vehicles. Peng et al.[12] train a recurrent network instead of a feedforward one and sample a different set of dynamics parameters for each rollout. Furthermore, with both visual and dynamics randomization, Open AI[6] trains a policy that manipulates the robot hand to rotate an object continuously to achieve the various target orientations.

However, Mehta et al.[13] indicates that UniformDR empirically leads to suboptimal policies that fail to generalize across multiple domains due to the uniform and broad sampling in the randomization space $\Xi$.

### 2.3.2 *Guided Domain Randomization*

Guided Domain Randomization (GuidedDR), a more sophisticated DR method, is developed to save the computation resources and avoid the redundant training cases that never occur in the real world. The intuition is substituting uniform randomization with guidance from task performance, real-world data, or simulator[5]:

#### 2.3.2.1 **Performance-guided**

Initially, we train a cluster of policies in the simulator varied with different randomization parameters. Then, we task these policies in the real world respectively and examine their performance. Hereinafter, $\xi_i$ is a set of randomization parameters and conforms to $P_\phi(\xi)$ in which $\phi$ is to be optimized by the performance of each trajectory $\tau_{real,\xi_i}$ that policy $\pi_{\xi_i}$ samples from the real world.

Cubuk et al.[14] have employed this idea to figure out the best data augmentation operations for image classification tasks in which the performance of each augmentation configuration is quantified as rewards for an RL policy. Similarly, Ruiz et al.[15] put forward the "learning to simulate" idea that uses meta-learning to find a distribution of simulated parameters that achieves the best performance in the target domain. The gist is to span the action space with all possible sets of randomization parameters $\xi$ and train a policy $\pi_\xi$ using Policy Gradient[16] to generate $\xi$ that maximizes the accumulative reward in the real tasks.

Muratore et al.[17] touch on a unique perspective: sample previously-unseen physics parameters and see how much worse the current policy performs on them than one trained on them so that we could adopt early-stopping to avoid overfitting to the simulator.

#### 2.3.2.2 **Reality-guided**

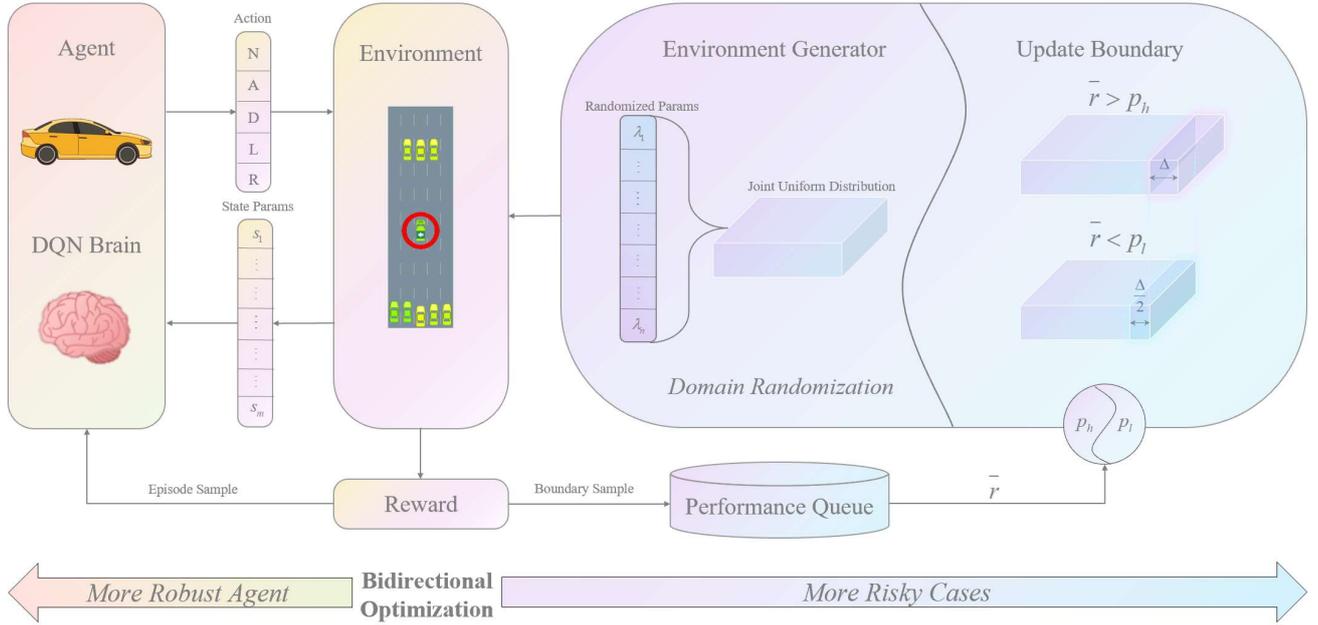

**Figure 1: Surface corner cases and robustify autonomous vehicle policies simultaneously via bidirectional optimization**

Reality-guided DR techniques are mostly to match the simulator with the real data distribution. It seems to be the combination of DR and SI / DA.

The SimOpt[18] model is trained in a randomized simulator and the policy is implemented to collect trajectories $\tau_\xi^{ob}$ and $\tau_{real}^{ob}$ in the simulation and real world. By minimizing $D(\tau_\xi^{ob}, \tau_{real}^{ob})$, a measure of discrepancy between the two kinds of trajectories, SimOpt model updates the randomization distribution $P_\phi(\xi)$ iteratively to incorporate the real-world data.

Instead of approaching the simulated data to the real distribution directly, the goal of Meta-Sim[19] is to first generate synthetic datasets of traffic scenarios that are usually naïve and unrealistic, and make them more physically possible with a little bit of real-world data.

Unlike the above-mentioned Reality-guided DR methods, what RCAN(short for Randomized-to-Canonical Adaptation Networks[20]) touches on is something different. To start with, a cGAN model is trained to generate non-randomized scenarios(namely canonical scenarios, the same hereinafter) for the randomized ones, and train RL policies in the canonical scenarios. Likewise, translate the real-world scenarios into the canonical counterparts with the same cGAN model so that policies are applied in a consistent environment for both simulated training and real-world validation. Undoubtedly, it performs a better transferrable ability compared to UniformDR.

### 2.3.2.3 Simulator-guided

Intuitively, the harder the simulated scenarios are than the real-world ones, the more transferrable our policies trained in these scenarios will be. Whereas, most of the examples in the simulator maybe not really that hard. The question is that if there are some automated ways of surfacing the hardest samples in the simulators so that we could focus the model training on the hardest ones. Without any real-world data for calibration or adaptation, simulator-guided DR is a pure domain randomization technique that is extremely appropriate for autonomous driving, circumventing the unsafe, laborious, and time-consuming real data collection part.

Active Domain Randomization(ADR)[13] provides two separate environments: randomized and reference(non-randomized) simulator. They have the intuition that the larger the discrepancy between the policy rollouts in these two simulators is, the harder samples the randomized simulator might have, the higher reward should be fed into Stein Variational Policy Gradient[21] particles so that these hard examples are surfaced and encouraged to explore.

More randomization leads to better transferability but an excessively wide randomization range often results in poor performance. To this end, Automatic Domain Randomization (AutoDR)[22] is framed to gradually widen the range if the performance is good near the boundary, and on the contrary, narrow the range. It is noteworthy that since the performance is collected in the simulator environment, AutoDR is not above-mentioned *performance-guided.* See more about AutoDR in the Appendices.

## 3 Problem Formulating

### 3.1 Methodology and Implementation

How to surface corner cases and robustify autonomous vehicle policies simultaneously is graphically formalized as a bidirectional optimization problem in Figure 1. Despite its similarity with the adversarial methods[23][24], this framework could step further by yielding a curriculum of scenarios from easy to difficult more targetedly.

　　　Regarded as a Visual Domain Randomization problem in this research, the variational simulation environment is structured by a set of Randomized Parameters $\lambda = [\lambda_1, \lambda_2, \cdots, \lambda_n]$. Corresponding to what [22] puts forward, each component of the parameter vector $\lambda_i$ complies with a uniform distribution $P_{\Phi_i}$, bounded by $\phi_i^L$ and $\phi_i^H$. Thus,

$$\lambda_i \sim P_{\Phi_i} = U(\phi_i^L, \phi_i^H) \quad (2)$$

and $\lambda_i (i=1,\cdots,n)$ comply with a joint uniform distribution. Distinguished from the *Randomized Parameter* $\lambda_i$, $\Phi_i$ is named as *Distribution Parameter*.

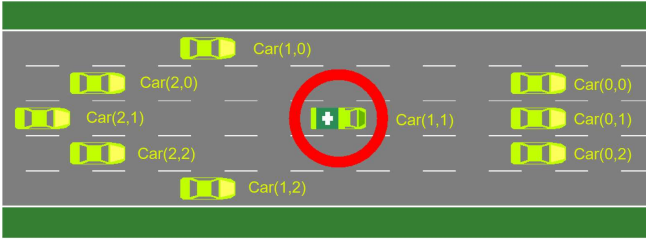

**Figure 2: Set up in a simple scenario**

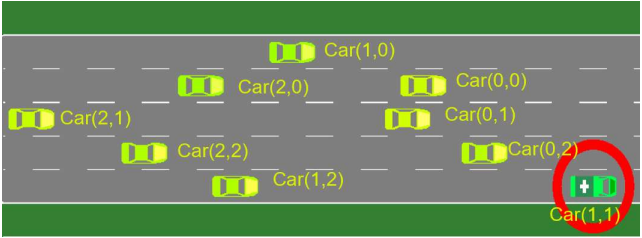

**Figure 3: Survive in a harder scenario**

　As is often the case, the reward value derived from the interaction between agent and environment counts towards the estimation value for the current observation(a set of state parameters) so that the agent could try to learn an optimal policy from it. This is merely a paradigm of monodirectional optimization. However, the reward value could also be redefined as the performance of the agent at a given state whereby the model could widen the distribution bound $[\phi_i^L, \phi_i^H]$ if the performance near the boundary of the distribution range is averagely beyond a threshold, and vice versa. According to the performance, this AutoDR technique produces a curriculum of harder tasks for the agent to adapt to. Combining AutoDR with the RL agent brings out this framework, Domain Randomization Reinforcement Learning(DR²L), a bidirectional optimization solution dedicated to the above-mentioned problem.

　　AutoDR plays the role of the environment generator in the bidirectional optimization solution in Figure 1. Beforehand, we need to claim that: Sampling each randomized parameter $\lambda_i$ uniformly from the range $[\phi_i^L, \phi_i^H]$ like UniformDR is called *Episode Sample*, while selecting one randomized parameter to lie on the boundary of the range and sampling the rest n-1 of them from the joint uniform distribution is named *Boundary Sample*. At each timestep in DR²L model (Figure 1), we do episode sampling with probability $p_b$ to train RL agent, while doing boundary sampling with probability $1 - p_b$ and collecting the performance value on each boundary into a queue buffer $D_i$ (corresponding to boundary $i$) with a certain capacity $N$. When $D_i$ gets filled with the performance value(i.e. current reward), average them and clear the queue buffer. If the average performance $\bar{r}_i$ is out of the threshold range, namely $\bar{r}_i > p_i^h$ or $\bar{r}_i < p_i^l$, correspondingly increase or decrease the distribution parameter $\phi_i$ to widen or narrow the range.

　　To optimize AutoDR in [22], our model for environment generating attempts from two perspectives:

- First and foremost, we halve the step length of decreasing $\phi_i$ against that of increasing $\phi_i$. Intuitively, if the average performance on the former boundary ($\phi_i^{t-1}$) is higher than $p_i^h$ while the one on the newly-expanded boundary ($\phi_i^t = \phi_i^{t-1} + \Delta$) is less than $p_i^l$, the downstream step is better to find the compromise ($\phi_i^{t+1} = \phi_i^t - \frac{\Delta}{2} = \phi_i^{t-1} + \frac{\Delta}{2}$) instead of stepping back ($\phi_i^{t+1} = \phi_i^t - \Delta = \phi_i^{t-1}$) meaninglessly. In this way, the boundary tends to converge at a more precise value so that our model could possibly find the hardest cases that autonomous vehicles could still adapt to in the end.

- What's more, giving up the fixed thresholds, we replace the $p_i^h$ or $p_i^l$ with the out-of-threshold $\bar{r}_i$ so that the model automatically tailors the thresholds to meet the transformation of environment and the learnability of the agent, leaving out the redundant manual tuning.

　　That is how the environment generator functions. To complement the bidirectional optimization loop depicted in Figure 1, reward function that measures performance in a canonical and well-founded way needs to be formulated. In this model, the design of reward function addresses two aspects: safety and efficiency, and the counterparts are respectively symbolized as $r_{safety}$ and $r_v$:

$$r_v = \frac{v - v_{max}/2}{v_{max}} \quad (3)$$

$$r_{safety} = r_{collision} \cdot I(collision) + r_{arrive} \cdot I(arrive) \quad (4)$$

In Equation 3 and 4, $v$ and $v_{max}$ are the current velocity and the maximum velocity of the ego car respectively. Additionally, $I(\cdot)$, known as the indicator function, is taken as 1 if the event $(\cdot)$ happens at the current timestep, and zero otherwise. To guarantee that cumulative reward could best measure the performance of the agent, we have to quantitatively define the constraints:

$$r_{arrive} + \inf \int_0^T r_v dt \geq r_{collision} + \sup \int_0^T r_v dt \quad (5)$$

Equation 5 eliminates the case that autonomous vehicles driving at the maximum velocity with collision happening cumulate higher reward than ones driving at the slowest velocity safely, where $T$ is the total traversal time. We assume that the slowest velocity is a positive number just greater than zero, and define the total traversal distance as $d$. Theoretically, Equation 5 could be simplified as follow:

$$r_{arrive} - r_{collision} \geq \frac{2d}{v_{max}} \quad (6)$$

As long as $r_{arrive}, r_{collision}, v_{max}$ and $d$ coincide with Equation 6, we could confirm that the reward design in Equation 3 and 4 are qualified to measure the performance. The current reward at timestep $t$ is summed up as:

$$R_t = r_v + r_{safety} \quad (7)$$

When we do boundary sampling at timestep $t$, the current reward is to be appended in the performance queue buffer $D_i$ associated with the selected boundary of the selected distribution parameter(boundary index: $i$):

$$D_i \leftarrow D_i \cup \{p \leftarrow R_t\} \quad (8)$$

### 3.2 Environment Setup

We conduct a series of experiments in SUMO[25] emulator and the preliminary scenario is set as Figure 2. The simulation environment incorporates a 5-lane highway, an ego car, and 8 neighboring cars. Initially, Car{(2, 0), (2, 1), (2, 2)} are longitudinally 20 meters behind the ego car Car(1, 1), and Car{(1, 0), (1, 2)} are longitudinally 10 meters behind Car(1, 1), and Car{(0, 0), (0, 1), (0, 2)} are longitudinally 20 meters in front of Car(1, 1). All the neighboring cars keep a constant velocity and are fixed on a certain lane. Besides, Car(1, 1), the DeepRL-based autonomous vehicle to be detailed in the next subsection, starts out at a distance of 200 meters away from the destination and could accelerate or decelerate as an autonomous agent in the whole journey.

Randomized parameters incorporate the initial velocity of each car, complying with a dynamic uniform distribution $\lambda_i(v_i) \sim U(\phi_i^L, \phi_i^H)$ where distribution parameters $\phi_i^L$ and $\phi_i^H$ are bounded by the minimum and maximum velocity. All the hyperparameters that the environment generator needs are manually tuned and listed in Table 1.

### 3.3 Agent Setup

Deep Q-Network, a.k.a. DQN[26], is one of the most practical deep reinforcement learning network architectures that have been widely implemented in autonomous driving research. Other than the design of reward function in *Methodology*, the state space and action space are another two crucial components for an RL agent. In this paper, the DQN agent refers to the ego car, Car(1,1) in Figure 2.

The state space entails not only the randomized parameters, the velocity of each car, but also the relative distances between neighboring cars and the ego car, structured in a flexible and data-efficient way[27]. The observation of the state space at timestep $t$ is $O_t$.

**Table 1: Hyperparameters of the Environment Generator**

| Hyperparameter | Value |
|---|---|
| Boundary sampling probability $p_b$ | 0.5 |
| Boundary update step length $\Delta$ | 0.5 |
| Initial increase threshold $p_h$ | 15.0 |
| Initial decrease threshold $p_l$ | 13.0 |
| Size of performance queue $N$ | 10 |
| Maximum value of $\phi_i^H$(m/s) | 30.0 |
| Minimum value of $\phi_i^L$(m/s) | 0.0 |

The action space accommodates five kinds of behaviors: accelerating longitudinally at 3m/s² in this timestep (A), decelerating longitudinally at 3m/s² in this timestep (D), making a left-lane-change (L), making a right-lane-change (R) and no-operation (N). The action chosen at timestep $t$ is $A_t$.

In terms of the network architecture, the target network incorporates an input layer, two hidden layers, and the output layer that calculates the Q-values for each action in the action space. The evaluate network is of the same structure as the target network but updates its parameters asynchronously: to lead to the final convergence, parameters of the target network $\theta$ are assigned to those of the evaluate network $\theta^-$ every $\mathcal{N}$ training iterations.

As for the learning process, action($A_t$) is selected by Q-values, otherwise, action will be chosen randomly with the probability of $\epsilon$ (exploration rate, decaying each step until reaching the minimum).

The interaction between the agent and the environment generates the next state $O_{t+1}$ and reward $R_t$. Every piece of transition $(O_t, A_t, R_t, O_{t+1})$ is stored in the memory pool with the capacity of $\aleph$ transitions. Randomly Selecting a mini-batch of transitions from the memory pool every training step, Experience Replay[28] technique helps DQN make full use of the experience and eliminate the correlation of sequential transitions. The ultimate goal is to minimize the loss function below:

$$L(\theta) = \left( R_t + \gamma \max_{A_{t+1}} Q(O_{t+1}, A_{t+1}; \theta^-) - Q(O_t, A_t; \theta) \right)^2 \quad (9)$$

All the hyperparameters of the DQN agent are manually tuned and shown in Table 2:

**Table 2: Hyperparameters of the DQN Brain**

| Hyperparameter | Value |
| --- | --- |
| Size of input layer | 22 |
| Size of hidden layer I | 20 |
| Size of hidden layer II | 10 |
| Exploration rate $\epsilon$ | 0.9→0.1 |
| Decaying rate of the exploration rate $\tau$ | 4×10$^{-6}$ |
| Updating the evaluation network every $\mathcal{N}$ iterations | 5000 |
| Size of the memory pool $\aleph$ | 2000 |
| Size of mini-batch | 32 |
| Learning rate $\eta$ | 0.001 |
| Discount factor $\gamma$ | 0.9 |

## 4  Evaluation and Analysis

In this paper, we put forward the DR$^2$L model to "surface harder cases" and "robustify autonomous vehicles in these cases" simultaneously. Therefore, evaluation and analysis from both perspectives are going to be touched on in this section.

### 4.1  Surfacing harder cases

During the training process, the model updates the distribution parameters $\phi_i^L$ and $\phi_i^H$ (the boundaries of the uniform distributions that environment parameters are sampled from) for many times. As mentioned before, the initial scenario like Figure 2 could gradually transfer to the harder ones like Figure 3. In Figure 4, the high-dimensional sampling environment parameters are visualized via t-SNE in a 3D system of coordinates and it depicts the transformation of the sampled parameters after every update.

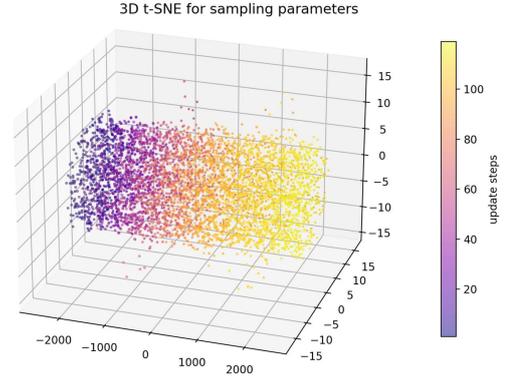

**Figure 4. Dimensionality reduction for sampling environment parameters via t-SNE**

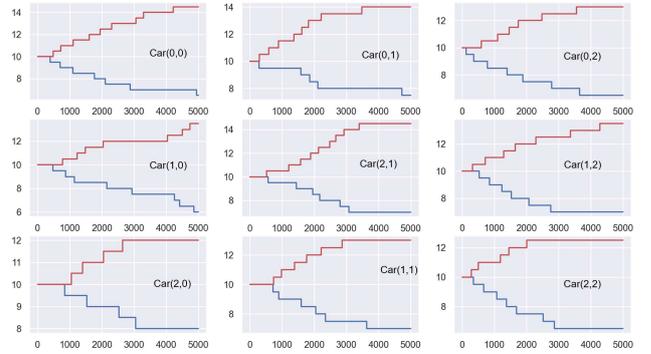

**Figure 5. distribution parameters updating dynamically during training**

In Figure 5, X-axis refers to the number of trained agents. Along the Y-axis, distribution parameters (the upper boundary and the lower boundary of each randomized parameter, namely the initial velocity of each car, are colored red and blue respectively) are dynamically changing during the training process, to gradually explore the boundary for harder cases. When most of the distribution parameters level off as Figure 5, we could believe the agents have discovered the hardest cases that they could just adapt to, indicating the balance between the agents and the environment generator achieved in this configuration.

In Figure 6, unlike the cumulative reward curve in RL tasks, it is visible that the cumulative reward in the DR$^2$L task has an abnormally large standard deviation at each smoothing value. Boundary samplings possibly account for the reward values that are significantly lower than the mean value because the DR$^2$L model is always surfacing harder cases for autonomous vehicles to train on. Also, the curve levels off, indicating the balance in the DR$^2$L model.

To summarize, Figure 4 and 5 indicate that DR$^2$L model is always generating variational distributions of randomized parameters to structure different environment cases, and Figure 6 reveals that these cases are always harder than those the agent could currently adapt to.

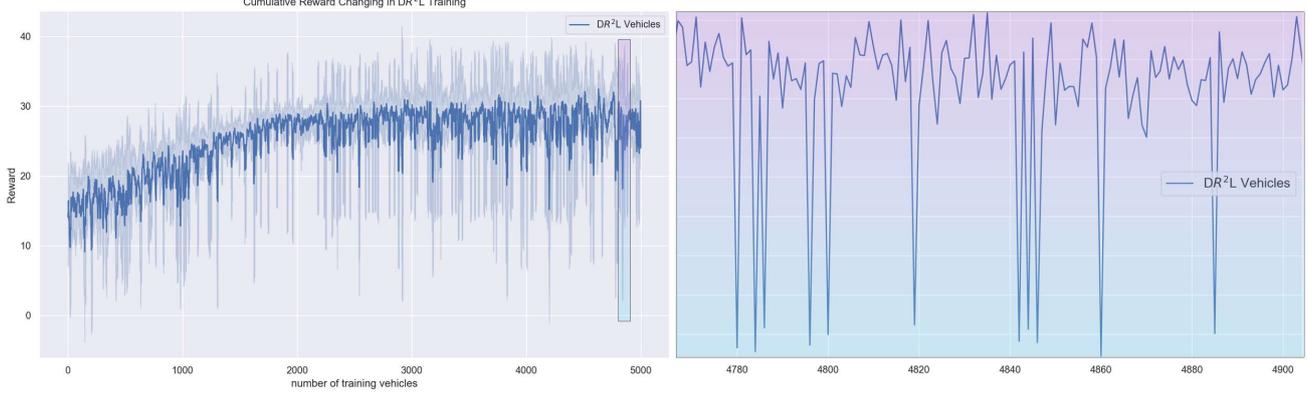

Figure 6: cumulative reward changing during the training with our DR²L model

Table 3: Evaluate the Performance of Agents Trained in Different Environments

| Test in / Train in | Easy Env | | Mid Env | | Hard Env | |
|---|---|---|---|---|---|---|
| | Avg Speed/ms$^{-1}$ | Collision-free | Avg Speed/ms$^{-1}$ | Collision-free | Avg Speed/ms$^{-1}$ | Collision-free |
| Easy Env | 9.95 | √ | 10.00 | × | 10.00 | × |
| Mid Env | 10.01 | √ | 9.31 | √ | 8.86 | × |
| Hard Env | 18.89 | √ | 19.07 | √ | 18.89 | √ |
| **DR Env(Ours)** | **21.07** | √ | **20.64** | √ | **19.82** | √ |

## 4.2 More Robust Agent

Other than the discussion in the previous subsection about Figure 6, it is noteworthy that the reward curve of the DR²L model also exhibits a tendency: firstly increase and then level off, which is similar to that in RL tasks. It corroborates that the agent is always learning to adapt to the harder cases provided continuously by the DR environment generator.

However, do agents trained in the DR²L model outperform those trained in environments with fixed distribution parameters? According to the information in Figure 5, we sample the distribution parameters when the number of trained agents is 0, 1000, and 4000, respectively corresponding to easy, mid, and hard environments. We reasonably believe that the hard environment provides almost the same cases as the ultimate ones, namely corner cases, generated by the DR²L model because most of the distribution parameters stop updating after 4000 agents have been trained as Figure 5 shows. Then, RL agents are trained in each environment with those fixed distribution parameters and compared with our DR²L model on safety and efficiency in these environments as Table 3. From the results came a suite of analyses:

- Firstly, the agents trained in environments with fixed distribution parameters are not capable of making safe decisions in harder environments;

- Besides, the agents trained in DR²L model drive not only faster but also more safely than the others do. It is inspiring to compare the performance of agents trained in the hard environment and the DR environment, which indicates that environments with dynamic distribution parameters are better than environments with fixed ones.

- To take these a step further, the DR²L model provides a curriculum of environments from the easiest to the hardest, in which the agents could learn more robust policies than they do in environments with fixed distribution parameters.

## 5 Conclusion and Future Work

In this work, we introduce the DR²L framework to figure out the bidirectional optimization problem: surfacing harder cases beyond the current ability of the agents and training the agents to adapt to them. The DR environment generator could produce a series of environments from easy to difficult, and finally to the corner cases, which refer to the hardest cases that the RL agents could adapt to. In summary, the RL agents trained in these environments with dynamic distribution parameters learn safer and more efficient policies.

Generally, Sim2real is a hierarchical concept, which could be better described as Sim2sim2real. Currently, we only concentrate on how to train agents in simulators for better transferability, referring to the Sim2sim step. For future research, we hope that the DR$^2$L method could realize the dream of bridging the ultimate Sim2real gap for DeepRL-based autonomous driving.

## ACKNOWLEDGMENTS

This work is supported by National Key R&D Program in China (2018YFB1600802), Tsinghua University Initiative Scientific Research Program (20183080016).

## APPENDICES

The algorithm of Automatic Domain Randomization mentioned in [22]:

```
Algorithm 1 ADR
Require: φ⁰                                          ▷ Initial parameter values
Require: {D_i^L, D_i^H}_{i=1}^d                      ▷ Performance data buffers
Require: m, t_L, t_H, where t_L < t_H                ▷ Thresholds
Require: Δ                                            ▷ Update step size
    φ ← φ⁰
    repeat
        λ ~ P_φ
        i ~ U{1,...,d}, x ~ U(0,1)
        if x < 0.5 then
            D_i ← D_i^L, λ_i ← φ_i^L                 ▷ Select the lower bound in "boundary sampling"
        else
            D_i ← D_i^H, λ_i ← φ_i^H                 ▷ Select the higher bound in "boundary sampling"
        end if
        p ← EVALUATEPERFORMANCE(λ)                   ▷ Collect model performance on environment parameterized by λ
        D_i ← D_i ∪ {p}                              ▷ Add performance to buffer for λ_i, which was boundary sampled
        if LENGTH(D_i) ≥ m then
            p̄ ← AVERAGE(D_i)
            CLEAR(D_i)
            if p̄ ≥ t_H then
                φ_i ← φ_i + Δ
            else if p̄ ≤ t_L then
                φ_i ← φ_i - Δ
            end if
        end if
    until training is complete
```

**Figure 7: AutoDR Algorithm**